\documentclass{article} 
\usepackage{nips13submit_e,times, hyperref, graphicx, amsmath, amssymb,amsthm,mathrsfs,amsfonts,dsfont}
\usepackage{url}

\title{Creating Scalable and Interactive Web Applications Using High Performance Latent Variable Models}

\author{
Aaron Q Li \\
Language Technologies Institute\\
School of Computer Science \\
Carnegie Mellon University\\
Pittsburgh, PA 15213 \\
\texttt{aaron@potatos.io} \\
\And
Yuntian Deng \\
Language Technologies Institute\\
School of Computer Science \\
Carnegie Mellon University\\
Pittsburgh, PA 15213 \\
\texttt{yuntiand@cs.cmu.edu} \\
\And
Kublai Jing \\
Language Technologies Institute\\
School of Computer Science \\
Carnegie Mellon University\\
Pittsburgh, PA 15213 \\
\texttt{hjing@andrew.cmu.edu} \\
\And
Joseph W Robinson\\
School of Computer Science \\
Carnegie Mellon University \\
Pittsburgh, PA 15213 \\
\texttt{jwrobins@andrew.cmu.edu} \\
}

\nipsfinalcopy 

\begin{document}

\maketitle
\begin{abstract}
In this project we outline a modularized, scalable system for comparing Amazon products in an interactive and informative way using efficient latent variable models and dynamic visualization. We demonstrate how our system can build on the structure and rich review information of Amazon products in order to provide a fast, multifaceted, and intuitive comparison. By providing a condensed per-topic comparison visualization to the user, we are able to display aggregate information from the entire set of reviews while providing an interface that is at least as compact as the ``most helpful reviews'' currently displayed by Amazon, yet far more informative.
\end{abstract}

\section{Introduction}
Latent variable models are a highly versatile tool for inferring the
structures and hierarchies from unstructured data. Typical use cases of latent
variable models include learning topics from documents, building user profiles,
predicting user behavior, and generating hierarchies of class labels.

Although latent variable models are arguably some of the most powerful tools in machine learning, they are often overlooked and underutilized in practice. Using latent variable models has many challenges in real world settings: models are slow to compute, hardly scalable, and results are noisy and hard to visualize. While prior work in \cite{LietalKDD2014,LietalWWW2015,SmoNar10,LietalNIPS2013,YaoMimMcC09,LietalANU2012} has addressed challenges in computational speed and scalability and systems such as \cite{NICTAOpinionWatchDiffGPU,ChaungetalCHI2012,SpheresVideoDiffGPU} have been proposed to visualize topic modeling results, there is yet a system that combines these components and presents the results to end users in an intuitive, effective, and meaningful way.

In this paper, we present the architecture of a modularized distributed system built on top of latent variable models that is able to process, analyze, and intuitively visualize large volumes of Amazon product reviews. We show a fully-featured demo using a completed prototype, and present the results with two case studies.

\section{Background}
\subsection{Current Amazon System}
The current Amazon system for presenting reviews to a potential customer consists of three distinct views: customer quotes, most helpful reviews, and the full review list. Below, we describe each of these in detail and outline the deficiencies that our system intends to resolve.
\subsubsection{Quotes}
\begin{figure}[ht!]
\centering
\includegraphics[width=0.8\textwidth]{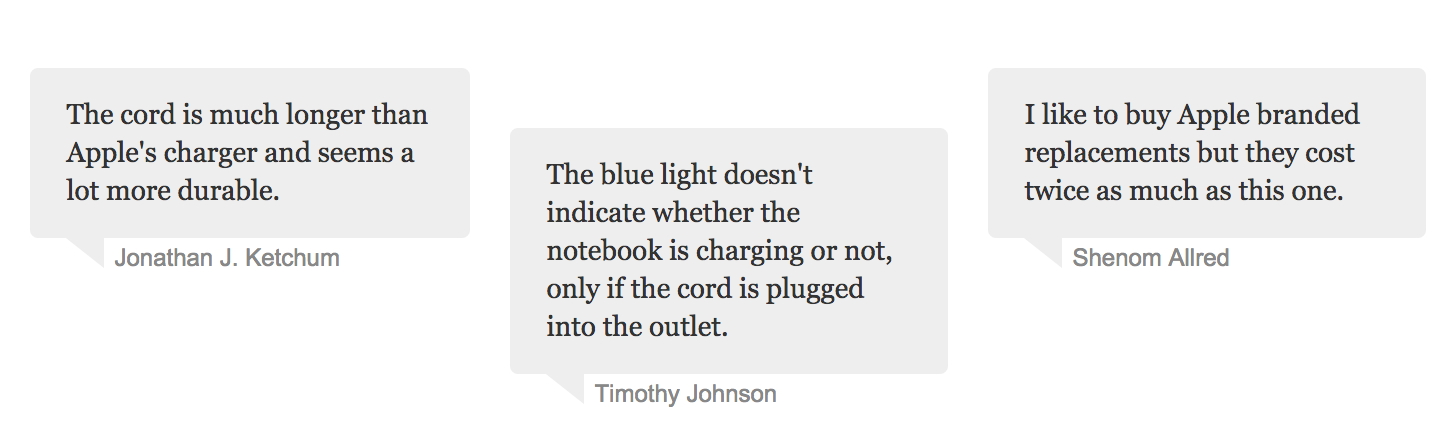}
\caption {Quotes for Macally PS-AC4 AC Power Adapter for Apple G4}
\label{figure_quotes}
\end{figure}
The quotes view provides a quick overview of a product and consists of three single-sentence quotations extracted from user reviews that are intended to capture sentiment about different aspects of a product in a concise and user-friendly way. As an example, the Macally Power Adapter quotes in Figure \ref{figure_quotes} display customer sentiment about the product's cord length and durability, a possibly annoying indicator light, and low cost relative to the brand-name alternative. While these quotes do provide insight into a product, a severe limitation is that they only represent the reviews of a few representatives. Regardless of how these representatives are chosen, their usefulness will always be limited by their inability to convey aggregate information on user sentiment towards different features of a product. We do, however, note the conciseness and simplicity of this representation. In designing our alternative system we attempt to retain as much of this simplicity as possible while improving information content.
\subsubsection{Most Helpful Reviews}
\begin{figure}[ht!]
\centering
\includegraphics[width=0.8\textwidth]{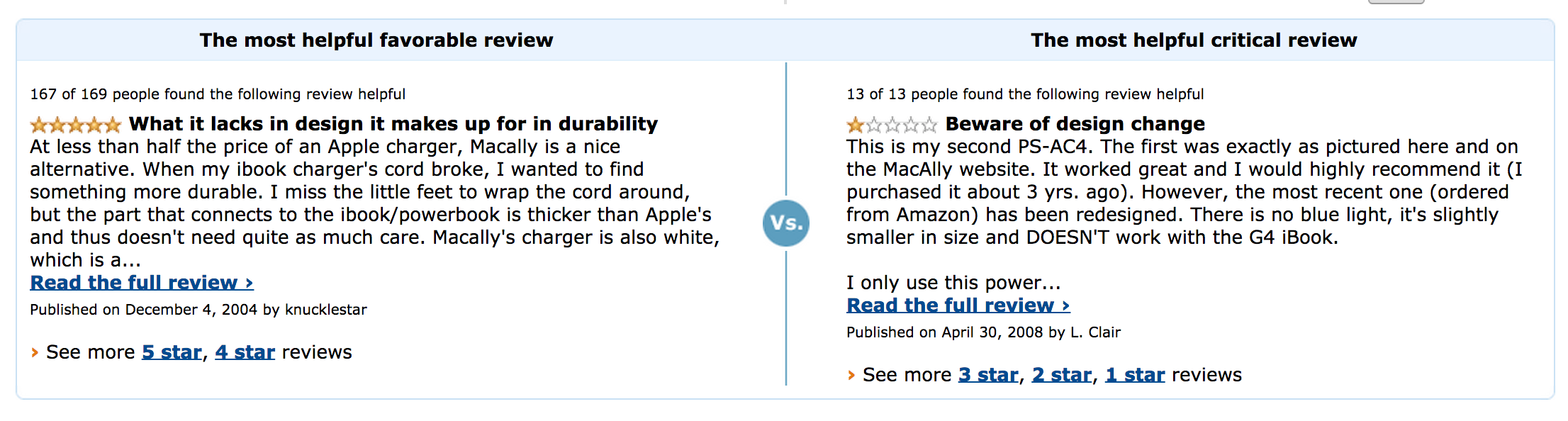}
\caption {Most helpful reviews for Macally PS-AC4 AC Power Adapter for Apple G4}
\label{figure_mosthelpful}
\end{figure}
The most helpful reviews section, displayed on the Amazon website above the full review list, provides the ``most helpful'' high and low rating reviews in a comparison-oriented view. An example of this is provided in Figure \ref{figure_mosthelpful}, again using the Macally power adapter. With this view, users can see two contrasting product experiences that have been deemed legitimate by other customers. One benefit of this is noise-reduction via user helpfulness feedback. However, these ``most helpful reviews'' can often be quite long and may only capture a few facets of the product. On top of this, helpfulness votes are strongly biased by a review's first vote -- a review that quickly gets a helpful vote is much more likely to get additional positive votes, and vice versa. In our system, we hope to address these problems by placing substantially less weight on helpfulness ratings when presenting information while simultaneously capturing a larger set of product facets and reducing the amount of textual content that needs to be read.

\subsubsection{Full Review List}
The full review list in helpfulness-sorted order is provided as a final source of reference. As there are hundreds of reviews for even moderately popular products, it is generally infeasible for a potential customer to examine this entire list. However, this view does provide the informational completeness that the previous two views lack. Our proposed system is designed to retain this completeness while simultaneously improving individual review search and allowing for multiple sorted review lists based on different aspects of a product that the user may be interested in.
\subsection{Latent Variable Models}

Many potentially suitable latent variable models exist for the task of analyzing Amazon reviews, such as unsupervised models described in \cite{XieXing2013, HuLiu2004, TitovMcDonald2008, BleNgJor03, Airoldi:2008:MMS:1390681.1442798}, supervised models in \cite{DiaoEtal2014,  BleiMcAuliffe2006}, and hierarchical models in \cite{Beal02theinfinite,SalHinton07,TehJorBeaBle06,CheBunDinXieDu12}. However, few of them have been designed with efficient inference and sampling procedures, such as \cite{TehNewWel2007, LietalKDD2014,LietalANU2012, journals/corr/abs-1212-2512,YaoMimMcC09,BunHut10,CheDuBun11}. For practical reasons, we only illustrate how our system works with latent variable models accompanied with efficient samplers, such as those introduced in \cite{LietalKDD2014, YaoMimMcC09}.
\subsubsection{Latent Dirichlet Allocation}

Our prototype makes use of Latent Dirichlet Allocation (LDA) \cite{BleNgJor03}, a widely used topic model in which one assumes that documents are generated from mixture distributions of language models associated with individual topics. That is,  the documents are generated by the latent variable model below:
\begin{figure}[ht!]
\centering
\includegraphics[scale=0.25]{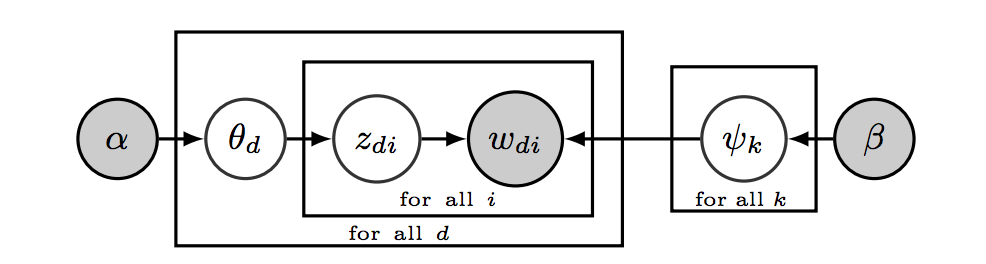}
\end{figure}
\\ For each document $d$ draw a topic distribution $\theta_d$ from a Dirichlet distribution with parameter $\alpha$
\begin{align}
\theta_d \sim Dir(\alpha)
\end{align}
For each topic $t$ draw a word distribution from a Dirichlet distribution with parameter $\beta$
\begin{align}
\psi_t \sim Dir(\beta)
\end{align}
For  each word $i \in \{1...n_d\}$ in document $d$ draw a topic from the multinomial $\theta_d$ via
\begin{align}
z_{di} \sim Mult(\theta_d)
\end{align}
Draw a word from the multinomial $\psi_{z_{di}}$ via
\begin{align}
w_{di} \sim Mult(\psi_{z_{di}})
\end{align}
The Dirichlet-multinomial design in this model makes it simple to do inference due to distribution conjugacy -- we can integrate out the multinomial parameters $\theta_d$ and $\psi_k$, thus allowing one to express $p(w,z|\alpha,\beta,n_d)$ in a closed-form \cite{YaoMimMcC09}. This yields a Gibbs sampler for drawing $p(z_{di}|rest)$ efficiently. The conditional probability is given by
\begin{align}
p(z_{di}|rest) \propto \frac{(n_{td}^{-di} + \alpha_t)(n_{tw}^{-di} + \beta_w)}{n_t^{-di} + \bar{\beta}}
\end{align}
Here the count variables $n_{td},n_{tw}$ and $n_t$ denote the number of occurrences of a particular (topic,document) and (topic,word) pair, or of a particular topic, respectively. Moreover, the superscript $.^{-di}$ denotes count when ignoring the pair $(z_{di},w_{di})$. For instance, $n_{tw}^{-di}$ is obtained when ignoring the (topic,word) combination at position $(d,i)$. Finally, $\bar{\beta}:=\sum_{w}\beta_w$ denotes the joint normalization.

Sampling from (5) requires $O(k)$ time since we have $k$ nonzero terms in a sum that need to be normalized. In large datasets where the number of topics may be large, this is computationally costly. However, there are many approaches for substantially accelerating sampling speed by exploiting the topic sparsity to reduce time complexity to $O(k_d + k_w)$ \cite{YaoMimMcC09} and further to $O(k_d)$ \cite{LietalKDD2014}, where $O(k_d)$ denotes the number of topics instantiated in a document and $O(k_w)$ denotes the number of topics instantiated for a word across all documents.

\section{Proposed System}
We propose a system that is capable of handling concurrent requests of product names from many users, analyzing relevant reviews by efficiently sampling from latent variable models, and displaying the streamlined updates from the models in real-time.

\subsection{Architecture}
\begin{figure}[!ht]
    \centering
    \includegraphics[width = 0.9\textwidth]{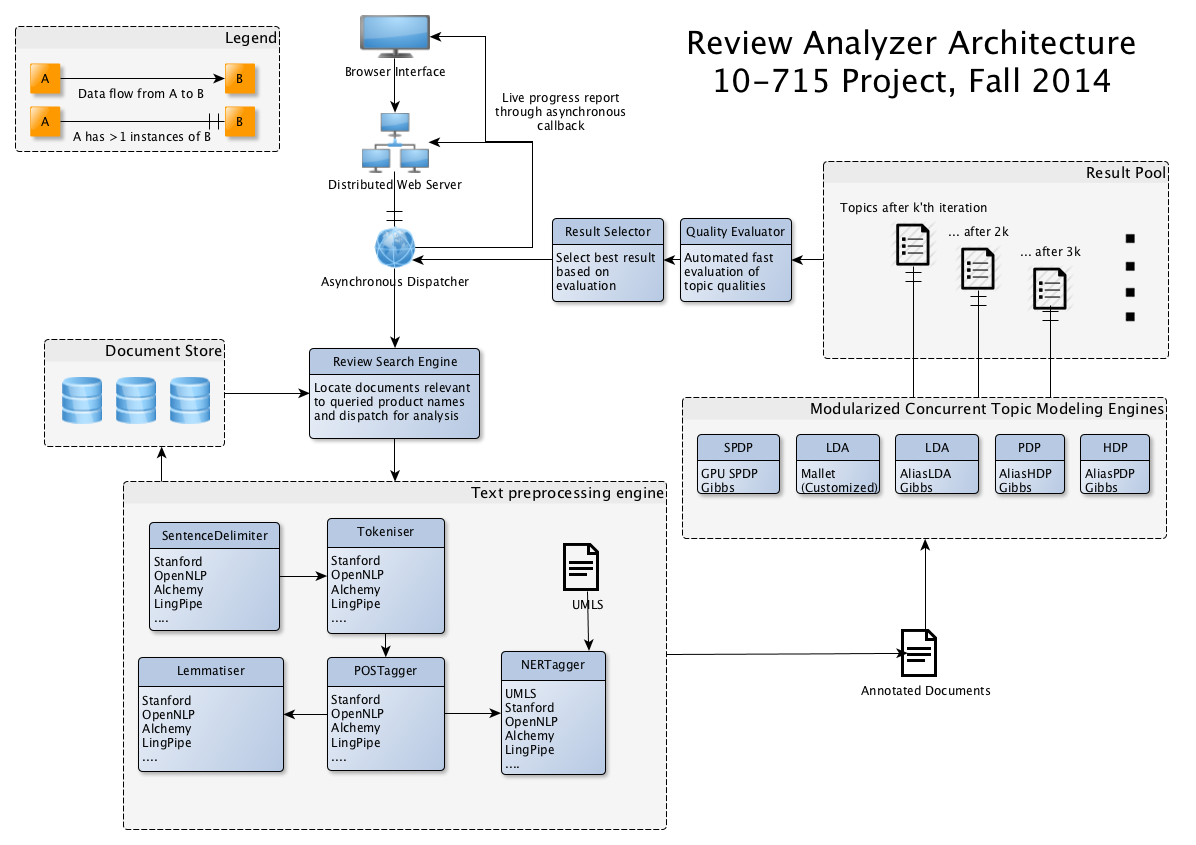}
    \caption{Architecture}
    \label{fig:arch}
\end{figure}
Figure \ref{fig:arch} shows the architectural design of our proposed system. In this design, almost every component is asynchronous and stateless. Furthermore, many tasks can be performed in a distributed fashion in order to balance computational costs.
\subsubsection{Components}
The major components of our system are the distributed web servers, asynchronous task dispatcher, data warehouse, search engine, text pre-processing engine, modularized concurrent topic modeling engine, update pool, and quality evaluator.

The workflow of the system is as follows: After a query is accepted by one of the web servers, a request is dispatched to the review search engine. The search engine determines if pre-processed results already exist in the data warehouse, and if not, how much further pre-processing is required. Requests of pre-processing tasks are created and dispatched to the text processing engine, where background workers are constantly pre-processing fresh reviews while assigning highest priority to new requests from the search engine. When pre-processed data is ready, the results are dispatched to the concurrent topic modeling engine. Multiple topic modeling instances are created at the same time, each emitting updates every few Gibbs sampling iterations to a pool where the results are further evaluated for quality. After evaluation, the best result is selected and sent back to the asynchronous task dispatcher, where it is later routed back to the initial web server. The web server then packages the result and returns it for presentation to the end user.

\subsubsection{Distributed Computing}
Our use of multiple samplers begs the following question: What is the best way to efficiently evaluate the quality of model results? The most popular two evaluation metrics, log likelihood and perplexity, both have their own merits. Compared to log likelihood, perplexity is typically regarded as a more accurate metric when the amount of test data is large, but is much slower to evaluate. In our system, we chose to use log-likelihood as the primary measure of quality for the following reasons:
\begin{itemize}
\item The number of reviews can be very limited for some products. Computing perplexity on insufficient test data may yield inaccurate evaluation of quality.
\item There is almost zero overhead for computing log-likelihood.
\end{itemize}
\subsection{Towards Large Scale}
The modularized and stateless design allows for easier scalability towards large scale processing. For example, Cassandra for data storage, Play! for distributed web servers, Parameter Server for topic modeling engines, and UIMA for pre-processing can all be swapped into our system in order to scale up with minimal modifications to other modules.
\subsection{Visual Design}
To visualize the thousands to millions of parameters estimated by latent variable models, we design an intuitive visualization framework that is most suitable for topic models. We analyze the meaning and structure of the statistical models, their sparsity properties and interconnections, and how they can be mapped onto a two dimensional Euclidean space. We additionally divide presentable information into a hierarchy in which a subset of information is available to the user at each level.

\section{Implementation}
Our prototype system uses the Electronics category from the SNAP Amazon reviews dataset \cite{amazondata}, a subset that consists of approximately 1 million reviews and 76,000 products. The primary components of our system -- pre-processing, database, modeling, and visualization -- are described below.
\subsection{Pre-Processing \& NLP}
Our pre-processing pipeline is built primarily on Stanford CoreNLP, a robust and highly modular package that has been shown to provide state-of-the-art performance on a variety of tasks. Using CoreNLP, we transform raw review text into a lemmatized token form augmented with part-of-speech. Ratings are used in place of sentiment to allow for a more intuitive per-topic rating and to reduce computation time. On our machine, we found that pre-processing time increased from 4.1ms to 706ms per review when sentiment analysis was added, making real-time sentiment computation with CoreNLP infeasible. Similarly, our original intension was to include named-entity recognition in our pre-processing pipeline. Our tests found that Stanford's implementation of this is far too slow -- on our machine, adding named-entity recognition resulted in an increase in average review pre-processing time from 4.1ms to 93.6ms.

While CoreNLP is used for base pre-processing, we defer stop word removal and instead filter out these words in real-time. This provides greater system flexibility and allows us to adjust stop word lists within seconds. From our tests, we found that filtering out tokens corresponding to words in a product's name resulted in substantially better and more visualizable topics. We also invested in augmenting MALLET's standard stop word list with many Amazon-specific stop words, such as ``review'' and ``star''.
\subsection{Database}
We use high memory instances on Google Cloud Engine (GCE) for our review data warehouse. NoSQL databases \cite{cattell2011scalable,chang2008bigtable} such as Cassandra \cite{lakshman2010cassandra} are ideal candidates for our setting as we require a high degree of concurrency and high availability, but relatively low demand for consistency. For the prototype system, a relational database is chosen to accomplish this task due to its simplicity and better performance on a relatively low volume of data. To reduce overhead, the same database is used as a cache for pre-processed data. A background task runs continuously to pre-process fresh reviews, giving priority to reviews of products being currently queried.

There are a few challenges in using a database for such a system. For instance, when a query is accepted, the system has to quickly determine if pre-processed reviews are available in the cache. If only part of the relevant reviews are cached, the fastest strategy to produce a reasonable response is to return the processed reviews and have the requester process the reviews not yet in the cache, then insert the processed results asynchronously as soon as they are available. When multiple queries are being processed concurrently, the chance that multiple on-demand pre-processing requests point to the same unprocessed review cannot be neglected. In this case, the same unprocessed review may be processed multiple times by different threads, incurring both computational overhead and potential risks of cache deadlock and duplication. One solution to this problem is to employ a scheduler to manage processing and updating, avoiding duplicate processing through the use of efficient hashing.
\subsection{Topic Modeling}
We use a customized version of the open-source MAchine Learning LanguagE Toolkit (MALLET) \cite{Mallet2002} as our modeling base. By generating a topic model from each product in a comparison in parallel, we achieve a significant speedup compared to MALLET's single threaded mode and substantially less overhead compared to MALLET's multithreaded mode. Our system is flexible with respect to the number of topics $k$ in that any number of topics can always be reduced to a core set for presentation to the user.

Since MALLET's implementation of LDA uses Gibbs sampling, we continue sampling until relative convergence. We empirically determine the number of iterations for producing good results. In addition, we perform alpha/beta optimization every 100 iterations. We estimate review-topic and word-topic distributions, which we then feed into our model summarization engine. Our modified version of MALLET allows us to compute and emit intermediate summaries after the 10th iteration of sampling and again every 2 seconds until sampling is complete. This gives the user immediate feedback and allows us to provide improved results as they become available.

From our topic model and reviews, we summarize each product with its topics and review summaries. Each topic structure contains a probability, list of augmented lemmas, rating, nearest topic and distance to that topic for the comparison product, and a representative review. Probabilities represent the overall topic distribution for the model generated from a product's reviews, which we use as our metric for topic relevance when presenting results to the end user. Each augmented lemma consists of the lemma text and both normalized and unnormalized word-topic weights. To compute nearest topics, we chose to use the Hellinger distance metric between normalized word-topic weights. Per-topic ratings $R_t$ are estimated as $\mathds{E}_{d \sim\theta_{\cdot,t}}[r(d)]$.\\
The top review $d^*$ for each topic $t$ is found by maximizing the following metric over $d$:
\begin{align}
\frac{\theta_{d,t}}{\| r(d) - R_t \| - \frac{1}{10}\mathds{1}(h^+(d) - h^-(d) > 0) + 1}
\end{align}
Here, $h^+(d)$, $h^-(d)$ refer to the helpfulness and unhelpfulness votes for review $d$, respectively. This metric was designed to address several problems that arise when using $\theta$ alone. Namely, we add a slight preference towards reviews that were voted to be helpful in order to avoid poor-quality representatives. We also bias our metric towards reviews in which the user rated the product similarly to the topic's aggregate rating, reducing user confusion and the frequency of unexpected results; a user browsing a 1-star topic does not expect to see a 5-star representative review praising the product. From our experiments, we found this metric to produce good results and have subjectively found it to offer more informative and intuitive representatives compared to the pure probability-based baseline. We also experimented with weighted n-gram topic visualization as an alternative to singular lemmas, however we found that single words resulted in simpler and more easily interpretable word clouds than n-grams.

As previously mentioned, our backend system also computes review summaries. The fields of this structure -- user id, profile name, helpful votes, unhelpful votes, rating, time, summary, and $\theta$ probabilities -- are self-explanatory. A key thing to note is our exclusion of the full review text. By doing this, we substantially reduce bandwidth costs and response time while allowing for later querying of individual reviews when requested.

\subsection{Visualization}
To achieve satisfactory performance, we made use of multiple rendering buffers and an advanced object management framework. The geometry of each data point is computed in real-time. Review contents are pulled from servers on-demand to minimize traffic. We additionally made use of many invisible techniques to minimize traffic, computation, and rendering efforts in order to maintain a smooth user experience.

\section{Results}
Our final prototype allows a user to query product reviews using an intuitive web interface and visualize initial results in under 3 seconds in the majority of instances. The interface allows for easy topic rating visualization with the ability to quickly select individual reviews. Selecting a topic brings up a side panel containing a topic-specific product comparison along with topic-sorted review summaries. Further details are available in the case studies below.
\subsection{Case Studies}
Below, we provide two product comparison case studies to demonstrate the utility and simplicity of our system. In the first, we compare the Macally PS-AC4 AC Power Adapter for Apple G4 to the Apple USB Power Adapter for iPod (White). We then repeat this analysis with a separate case consisting of two cameras, the Canon Digital Rebel XT 8MP Digital SLR Camera with EF-S 18-55mm f3.5-5.6 Lens (Black) and the Sony Cybershot DSC-T1 5MP Digital Camera with 3x Optical Zoom.
\subsubsection{Power Adapters}
\begin{figure}[ht!]
\centering
\includegraphics[width=0.8\textwidth]{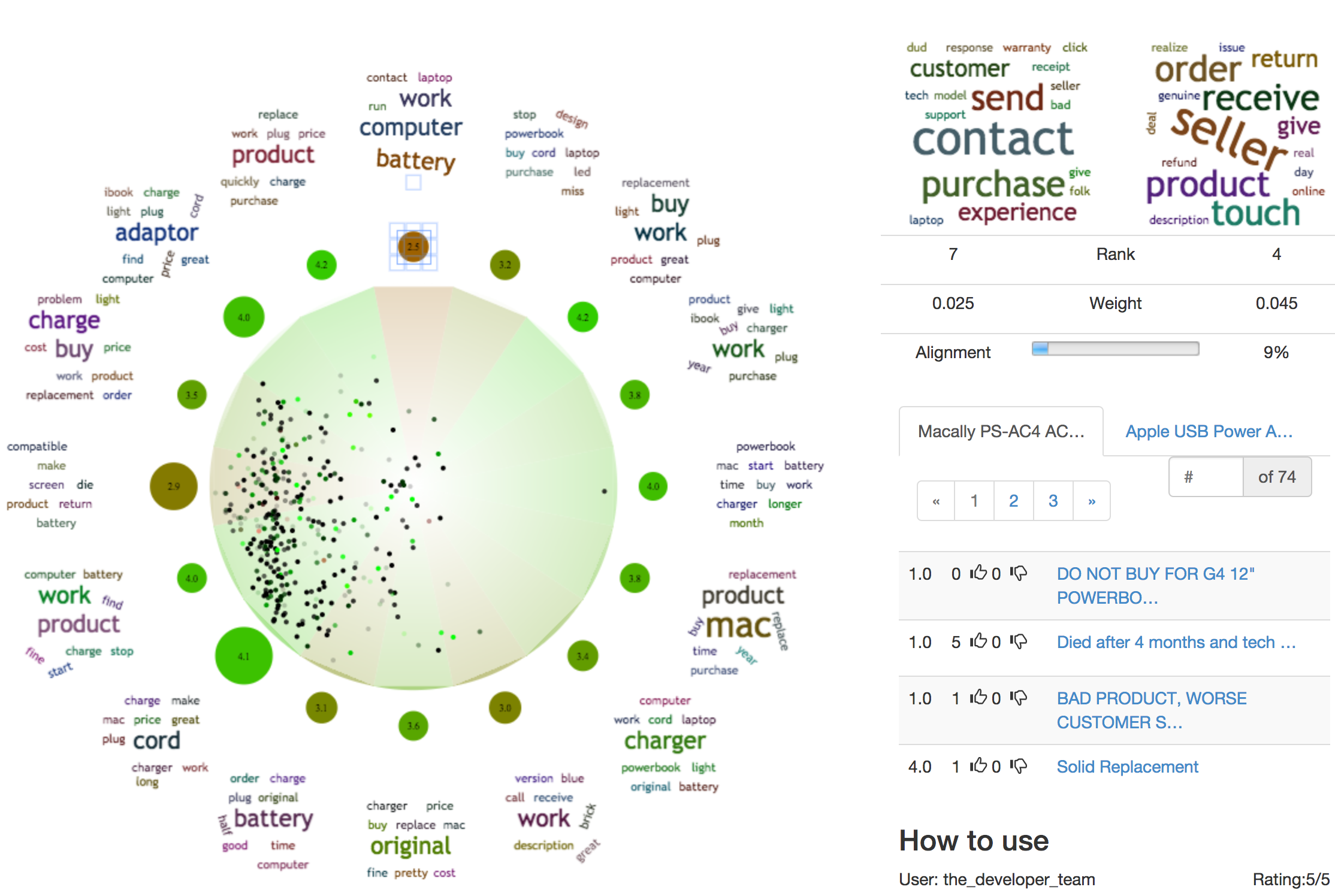}
\caption {Full interface view for a comparison of Macally PS-AC4 AC Power Adapter for Apple G4 and Apple USB Power Adapter for iPod (White)}
\label{figure_powercase}
\end{figure}
Figure \ref{figure_powercase} shows the interface a user of our system would see when comparing the two power adapters, using the Macally adapter as the reference product. On our machine, an initial view was available in approximately 2 seconds with updates continuing for a few seconds afterwards.

From the word clouds, we immediately see that customers tend to think highly of the Macally adapter's power cord, giving the topic high weight and an associated rating of 4.1. In contrast, we see 2.5 stars attributed to a topic dominated by ``work'', ``computer'', and ``battery''. Clicking on this topic circle brings up the panel on the right, where we quickly learn that this product has a tendency to die after a few months and intermittently not work, failing to charge the the customer's computer.
\subsubsection{Cameras}
\begin{figure}[ht!]
\centering
\includegraphics[width = 0.4\textwidth]{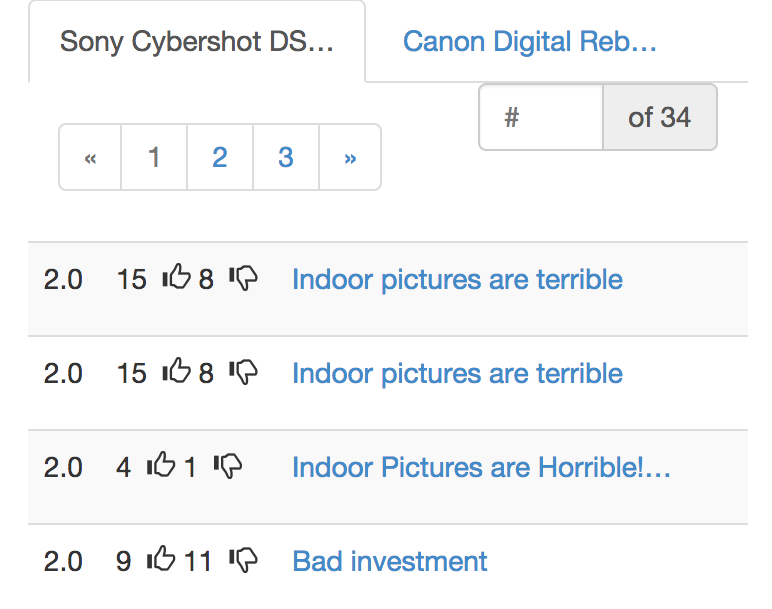}
\includegraphics[width = 0.4\textwidth]{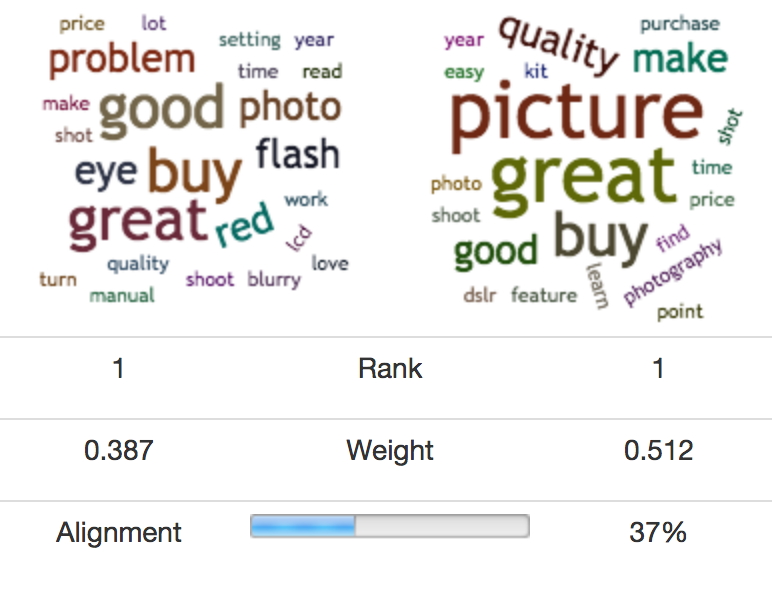}
\caption {Side panel (split) for a comparison of Sony Cybershot DSC-T1 5MP Digital Camera with 3x Optical Zoom and Canon Digital Rebel XT 8MP Digital SLR Camera with EF-S 18-55mm f3.5-5.6 Lens (Black)}
\label{figure_cameracase}
\end{figure}
Figure \ref{figure_cameracase} shows the side panel displayed for a single topic in the Sony-Canon camera comparison. As we can see, the Sony camera (left column) has potential red-eye and flash problems, whereas the nearest topic from the Canon model suggests higher design and picture quality. The relatively high similarity (37\%) of these topics suggests that the two do indeed describe the same facet of the cameras. Looking at the top topic reviews for the Sony camera, we see concerns about indoor picture quality compared to the Canon. As such, users whose primary concern is picture quality can quickly see that the Canon is generally regarded as superior in this area. However, cost conscious users that require a cheaper and more compact device can also benefit from this interface by learning that the Sony camera's problems tend to stem from picture quality issues rather than device failures.
\section{Future Work}
For future work, we wish to improve the scalability and performance of our system. In terms of modeling performance, replacing MALLET with higher performing implementations based on ideas described in \cite{LietalKDD2014, LietalANU2012, LietalWWW2015} would reduce sampling time and allow our system to scale to a larger number of users. We also expect future project iterations to use a continuously updating, crawler-based system in order to provide more up-to-date information and allow us to pre-process new data as it occurs.

We also plan to explore alternative topic distance metrics. Compared to our limited Hellinger distance metric, we suspect that including part-of-speech information will lead to better pairings. Specifically, we plan to experiment with computing distances on nouns only, allowing sentiment (typically manifested in adjective use) to arise naturally. Related part-of-speech experiments may also prove useful.

A known deficiency of topic models is the presence of noisy topics. Paralleling our use of per-topic customer ratings, we plan to explore per-topic helpfulness ratings based on our intuition that noisy and less informative topics would receive a higher proportion of weighted unhelpful votes and can therefore be automatically pruned out.

\newpage
\bibliographystyle{abbrv}

\end{document}